\documentclass{article}

\PassOptionsToPackage{numbers, compress}{natbib}
\usepackage[preprint]{neurips_2024}

\usepackage{wrapfig}

\usepackage{times}
\usepackage{soul}
\usepackage{url}
\usepackage[hidelinks]{hyperref}
\usepackage[utf8]{inputenc}
\usepackage[small]{caption}
\usepackage{color}
\usepackage{graphicx}
\usepackage{amsmath}
\usepackage{amsthm}
\usepackage{booktabs}
\usepackage{algorithm}
\usepackage{colortbl}
\usepackage{xcolor}
\usepackage{algorithmic}
\usepackage{amssymb}

\usepackage{multirow}
\usepackage{inconsolata}
\usepackage{xspace}

\newcommand{\xxx}{AgentCoder\xspace}
\usepackage{soul}
\usepackage[capitalize]{cleveref}
\crefname{section}{Sec.}{Secs.}
\Crefname{section}{Section}{Sections}
\Crefname{table}{Table}{Tables}
\crefname{table}{Tab.}{Tabs.}

\title{AgentCoder: Multi-Agent Code Generation with Effective Testing and Self-optimisation}

\author{%
  Dong Huang \\
University of Hong Kong\\
  \texttt{dhuang@cs.hku.hk} \\
  \And
  Jie M.Zhang \\
  jie.zhang@kcl.ac.uk \\
  \texttt{jie.zhang@kcl.ac.uk} \\
  \And
  Michael Luck \\
  University of Sussex \\
  \texttt{Michael.Luck@sussex.ac.uk} \\
  \And
  Qingwen BU \\
  Shanghai Jiao Tong University \\
  \texttt{qwbu01@sjtu.edu.cn} \\
  \And
  Yuhao Qing \\
  University of Hong Kong \\
  \texttt{yhqing@cs.hku.hk} \\
  \And
  Heming Cui \\
  University of Hong Kong \\
  \texttt{heming@cs.hku.hk} \\
}

\begin{document}

\maketitle

\begin{abstract}
Advances in natural language processing (NLP) have been significantly boosted by the development of transformer-based large language models (LLMs). These models have revolutionized NLP tasks, particularly in code generation, aiding developers in creating software with enhanced efficiency. Despite their advances, challenges remain in balancing code snippet generation with effective test cases. To address these issues, this paper introduces \xxx, a novel code generation solution comprising a multi-agent framework with a specialized test designer agents in addition to the programmer agent and the test executor agent. During the coding procedure, the test designer agent generates effective test cases for the generated code, and the test executor agent runs the code with the test cases and writes feedback to the programmer agent for it to refine the code. This collaborative system enhances code generation efficiency with less cost, outperforming both single-agent models and earlier multi-agent strategies, demonstrated by our extensive experiments on 14 LLMs and 16 baseline approaches. For example, \xxx (GPT-4) achieves 96.3\% and 91.8\% pass@1 in HumanEval and MBPP datasets with an overall token overhead of 56.9K and 66.3K, while state-of-the-art obtains only 90.2\% and 78.9\% pass@1 with an overall token overhead of 138.2K and 206.5K. 
\end{abstract}

\section{Introduction}\label{sec:intro}

In recent years, natural language processing (NLP) has been dramatically transformed by transformer-based large language models (LLMs). These models, notably exemplified by the GPT-x series~\cite{brown2020language,OpenAI2023GPT4TR} developed by OpenAI, have consistently set the benchmark for performance across a wide array of standard NLP tasks. One of the most pivotal applications for these LLMs is code generation for downstream tasks, where they play a vital role in aiding developers in creating software~\cite{codebert,wang2021codet5,wang2023codet5+,nijkamp2022codegen,nijkamp2023codegen2,Li2023StarCoderMT}. Through extensive pretraining on substantial code-related datasets, such as publicly available data on GitHub, these code LLMs acquire intricate contextual understanding that can be effectively applied to diverse code-related tasks.

Numerous recent efforts have been made to improve the effectiveness of LLMs by incorporating in-context learning and its variations~\cite{Dong2023SelfcollaborationCG,Wei2022ChainOT,Le2023CodeChainTM,Huang2023CodeCoTAB,Zhang2023SelfEditFC,Chen2023TeachingLL,Madaan2023SelfRefineIR}, where an important optimisation path is single-agent self-refinement within the same conversation. For example, \citet{Zhang2023SelfEditFC} proposed Self-Edit to enhance the performance of LLMs in code generation. In particular, Self-Edit runs the  generated code against test cases that are manually written by developers. It then prompts the LLMs to refine the code based on the error messages of failed tests. 
\citet{Huang2023CodeCoTAB} introduced CodeCoT, which uses LLMs to generate both code and test cases, thereby avoiding the reliance on developers for providing tests.

Recently, several studies (e.g., MetaGPT \cite{hong2023metagpt}, ChatDev \cite{qian2023communicative}, and AgentVerse \cite{Chen2023AgentVerseFM}) have proposed to use multi-agent collaborations to enhance the effectiveness of LLM-based code generation, where each agent addresses a unique task such as code generation or task planning. These multi-agent collaboration frameworks aim to overcome the limitations of single-agent methods by distributing the workload and optimizing performance across various aspects of the code generation process.
Nevertheless, these methods have two limitations: 1) they have less effective feedback mechanism to provide the LLMs with valuable information. For example, the accuracy of the generated tests from MetaGPT \cite{hong2023metagpt} is only 80\% for HumanEval;
2) they involve an excessive number of agents (e.g., MetaGPT has 5 agents, ChatDev has 7 agents), which require significant token resources for communication and coordination among different agents.

To address the above-mentioned challenge, in this paper, we propose \xxx,
a multi-agent code generation framework with effective test generation and small token overhead. 
\xxx has only three simple agents, i.e., the programmer agent, the test designer agent, and the test executor agent. The programmer agent interacts with advanced code generation models to create code based on coding requirements. The test designer agent designs accurate, diverse, and comprehensive test cases with code generation models independently based on the coding requirements. The test executor agent interacts with both the programmer agent and the test designer agent: it executes the tests from the test designer agent against the code generated by the programmer agent and then provides test execution results to the programmer agent. Once the feedback is obtained by the test executor agent from the local environment~(i.e., local terminal), it checks whether the feedback contains error information~(e.g., runtime error and assertion error). If all test cases pass the generated code, the test executor agent provides the code snippets to the human developer. Otherwise, the test executor agent feeds back to the programmer agent and then requires it to fix the bug reported in the feedback. The iteration then continues once the feedback is that all test cases pass the code snippets or the iteration budget is done.

The test executor agent plays a pivotal role by designing effective tests to critically evaluate the code.
Compared to existing test generation methods such as those used by CodeCoT and MetaGPT, \xxx has three unique features.
First, \xxx generates tests without seeing the whole code snippet, because the tests generated immediately following the code in one conversation can be biased and affected by the code, losing objectivity and diversity in the testing~(See \cref{tab:coverage}). 
Second, \xxx proposes generating tests independently from the source code generation, intentionally separating the code generation and test generation processes.
This choice is made based on previous findings that as the model achieves high performance in generating code snippets, there may be a corresponding decrease in the effectiveness of test case generation \cite{Chen2023AgentVerseFM,Zhang2023ProAgentBP}. This trade-off scenario occurs due to the model's limited resources and its focus on optimizing one aspect of the code generation process, which might inadvertently compromise the quality of other tasks~\cite{Chen2023AgentVerseFM,Zhang2023ProAgentBP}.
Third, the test designer agent in \xxx is carefully designed and prompted to generate basic, edge,
and large scale tests, yielding high accuracy and test coverage.

Our extensive experiments with 14 LLMs and 16 optimisation baselines demonstrate that \xxx significantly improves the effectiveness and efficiency of code generation, outperforming all baseline approaches. 
In particular, \xxx obtains an average of 91.5\% and 84.1\% pass@1 on all the datasets with GPT-4 and GPT-3.5, respectively, while the state-of-the-art obtains 86.8\% and 75.3\%.
The overall token overhead for AgentCoder is 56.9K for HumanEval and 66.3K for MBPP, significantly lower than other state-of-the-art muti-agent frameworks including MetaGPT (138.2K / 206.5K respectively),
ChatDev (183.7K / 259.3K), and AgentVerse (149.2K / 193.6K). 
Moreover, our test designer agent achieves a test generation accuracy of 89.6\% and 91.4\% for for HumanEval and MBPP with GPT-4, respectively, outperforming the second-best method MetaGPT whose accuracy is 79.3\% and 84.4\%. In terms of code coverage, our test designer agent achieves a line coverage of 91.7\% for HumanEval and 92.3\% for MBPP with GPT-4, while the coverage for MetaGPT is 81.7\% and 80.5\%, respectively.

Our main contributions are as follows:

\begin{itemize}
\item We propose \xxx, a multi-agent framework for code generation with effective test generation and small token overhead. \xxx contains three distinct agents, i.e., the programmer agent, the test designer agent, and the test executor agent.

\item We conduct an extensive evaluation with 14 LLMs and 16 LLM-based optimisation approaches which demonstrates that \xxx outperforms all the baselines in code generation. In particular, \xxx obtains 77.4\% and 89.1\% pass@1 with GPT-3.5, while state-of-the-art obtains only 69.5\% and 63.0\%. 

\item We conduct a deep analysis of our results and ablation studies, which demonstrate the contribution of different agents, the effectiveness of the tests generated by the test designer agent, and the necessity of using separate agents for code generation and test case design. 

\end{itemize}

\section{Related Work}

\subsection{Large Language Model for Code Generation}
Various architectures have been explored in these models, some notable examples being CodeBERT~\cite{codebert}, PLBART \cite{Ahmad2021UnifiedPF}, and CodeGPT~\cite{CERT}. These models are pre-trained on code corpora to develop a deep understanding of code syntax, semantics, and idiomatic constructs. Some innovative approaches integrate structured representations to enhance their comprehension of the complexities in code. For example, GraphCodeBERT~\cite{Guo2020GraphCodeBERTPC} incorporates graph-based representations, while CodeT5+~\cite{wang2023codet5+} combines the encoder-decoder paradigm with the structural essence of code. These enhancements aim to give the models a more fine-grained understanding of code relationships and dependencies beyond just syntactic patterns. A current trend is the construction of large scale models~(e.g.,  Codex~\cite{chen2021evaluating} and CodeGen~\cite{nijkamp2022codegen}) with billions of parameters, which have illustrated the performance of state-of-the-art in code generation tasks. Recently, foundation models~(e.g., GPT-3.5-turbo, GPT-4) have also been used for code generations~\cite{Madaan2023SelfRefineIR,Huang2023CodeCoTAB}. These foundation models illustrated the state-of-the-art performance for code generation tasks.

\subsection{Enhancing Code Generation through Prompt Engineering}
Recent advances in code generation have been significantly influenced by the integration of few-shot learning techniques with LLMs. A notable contribution in this realm is the concept of self-refinement with few-shot prompting, as proposed by \citet{Madaan2023SelfRefineIR}. This approach involves an LLM iteratively refining its own generated code, leading to significant improvement in code quality. Another approach is the Self-Debugging technique introduced by \citet{Chen2023TeachingLL}, which involves testing the generated code against user-provided test cases. In scenarios where such test cases are unavailable, the model engages in direct debugging by explaining the code, thus addressing potential issues. Complementing these methods, \citet{Huang2023CodeCoTAB} introduced CodeCoT, employing a Self-Exam Chain of Thought (CoT) process. This technique guides the model to generate code alongside test cases, particularly useful when external test cases are not available. CodeCoT adds a layer of logical reasoning to the code generation process. However, it is important to note that while this method can identify syntax errors, functional errors may still go undetected as both the code and its test cases are generated by the same model. Building upon these concepts, \citet{Dong2023SelfcollaborationCG} proposed the Self-Collaboration model, which divides the LLMs into different roles: an analyst, a coder, and a tester. The tester is powered by an LLM which predicts whether the code is buggy. Such practice may ignore many bugs in the code because the code is not executed in the local environments.

\subsection{Multi-agent Collaboration}
In recent months, LLM-based multi-agent frameworks have gained significant attention from both industry and academia. These frameworks can be broadly categorized into two groups: non-code generation and code generation multi-agent frameworks. Non-code generation multi-agent frameworks have been explored in various contexts. For example, Stable-Alignment \cite{liu2023training} generates instruction datasets by establishing consensus on value judgments through interactions among LLM agents in a sandbox environment. Generative Agents \cite{park2023generative} simulate a ``town'' of 25 agents to investigate language interaction, social understanding, and collective memory. NLSOM \cite{zhuge2023mindstorms} employs agents with different functions to solve complex tasks through multiple rounds of ``mindstorms''. \citet{cai2023large} propose a model for cost reduction by combining large models as tool makers and small models as tool users. Other works focus on cooperation and competition in planning and strategy \cite{meta2022human} or propose LLM-based economies \cite{zhuge2023mindstorms}. While these works have made significant contributions to non-code generation tasks, AgentCoder, specifically addresses code generation tasks, presenting unique challenges and opportunities for multi-agent collaboration in software development.

Several code generation multi-agent frameworks \cite{hong2023metagpt,li2023camel,Chen2023AgentVerseFM,qian2023communicative} have been proposed concurrently with \xxx in recent months. For example, MetaGPT \cite{hong2023metagpt} simulates the software development life cycle using multiple agents. 
However, these frameworks often face two significant challenges. First, they may have less effective feedback mechanisms to provide the LLMs with valuable information. For example, the accuracy of the generated tests from MetaGPT \cite{hong2023metagpt} is only 79\% for HumanEval, which limits the effectiveness of the feedback provided to the code generation agents. Second, these frameworks often require an excessive number of agents (e.g., MetaGPT has 5 agents, ChatDev has 7 agents), which can lead to significant token overhead for communication and coordination among different agents.
Different from these multi-agent frameworks, \xxx addresses these challenges by introducing a more efficient and effective approach. First, \xxx employs a dedicated test designer agent that generates accurate, diverse, and comprehensive test cases independently of the code generation process, ensuring the objectivity and effectiveness of the generated tests. Second, \xxx streamlines the multi-agent collaboration by utilizing only three agents: the programmer agent, the test designer agent, and the test executor agent. This design choice significantly reduces the token overhead associated with communication and coordination among agents, while still leveraging the benefits of multi-agent collaboration.

\begin{figure*}
    \centering
    \includegraphics[width=0.95\textwidth]{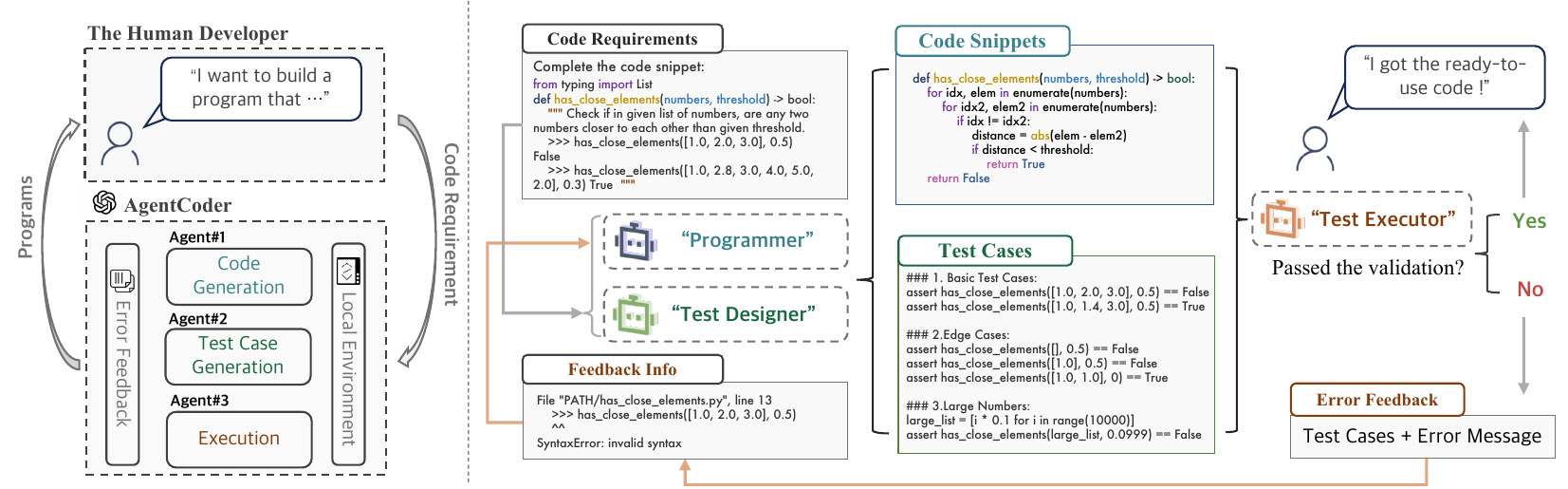}
    \vspace{-0.2cm}
    \caption{Pipeline of \xxx with a code generation example from HumanEval}
    \label{fig:pipeline}
    \vspace{-0.5cm}
\end{figure*}
\section{Methodology}

The framework of \xxx and its pipeline are illustrated in~\cref{fig:pipeline}. The process begins by inputting tasks/code generation requirements/descriptions into the code generation agent (Agent\#1: the programmer agent). 
Subsequently, the test case generator (Agent\#2: the test designer agent) is tasked with generating test cases, which are used to evaluate the correctness of the code snippets produced by the programmer agent. 
The code snippets and test cases are collected by the test executor agent (Agent\#3) and executed in the local environment~(local terminal) to obtain feedback~(i.e., whether the code passes all tests and the error message if the code fails for some tests). If the test executor agent finds that the code snippets pass all test cases, it will return the code to the user and finish the iteration. 
Otherwise, the test executor agent will return the test execution error messages to the programmer agent.
The iteration then continues, with the programmer agent regenerating code snippets to address the issues identified in the feedback,
and the test executor agent re-executes the new code and provides new feedback to the programmer agent, until the test executor agent finds that the code passes all the tests.

\subsection{Programmer agent: code generation with Chain-of-Thought instruction}
In our framework, The programmer agent is powered by LLMs. 
It needs to consider two scenarios, i.e., code generation and code refinement. Specifically, as shown in~\cref{fig:pipeline}, during the code generation stage, the human developer will require the programmer agent to generate code snippets to complete specific tasks, the programmer agent employs a Chain-of-Thought approach to simulate the typical programming process, methodically breaking down the task into smaller, manageable steps. 
The Chain-of-Thought process is instructed to contain four steps, i.e., problem understanding and clarification, algorithm and method selection, pseudocode creation, and code generation (the prompt and response example is shown in Appendix A.3 Figure 6 and 7). 

Taking the coding task \textit{Check if in given list of numbers, are any two numbers closer to each other than given threshold} (shown in~\cref{fig:pipeline}) as an example, during the initial code generation, the programmer agent will try to understand and clarify the given task, in this case interpreting the requirement to identify pairs of numbers in a list that are within a specified threshold of each other. The programmer agent will then decide on an algorithm or method to solve the problem. This could involve choosing an efficient way to compare each pair of numbers in the list. Next, during the pseudocode creation, the programmer agent will develop a step-by-step guide or pseudocode for the solution, ensuring a logical flow of operations. Finally, in the code generation stage, the programmer will translate the pseudocode into executable code.

Code snippets generated by the programmer agent can be incorrect, containing various types of errors (e.g., syntax and runtime errors), leading to failed test cases provided by the test designer agent.
Under such circumstances, the programmer agent will take feedback from other agents and refine the code snippets. The refinement process is iterative, with the programmer agent continuously enhancing the code based on feedback until the code successfully passes all test cases.

\subsection{Test designer agent: generating basic, edge, and large scale tests}

The test designer agent is also powered by LLMs. It is a crucial component of our \xxx's framework to test the code and provide reliable feedback for the programmer agent to optimise the code iteratively.   
We carefully designed the prompts for the test designer agent to satisfy the following three expectations:
(i) to generate basic test cases, (ii) to cover edge test cases, and (iii) to cover large scale inputs (the test designer agent's prompt and response example is shown in Appendix Figure 8 and 9).
The first aspect expects that the test designer agent designs test cases that cover the fundamental functionality of the code. These tests are designed to ensure that the code performs as expected under normal conditions. For instance, in a task that involves sorting a list, the basic test cases  verify that the list is sorted correctly for typical inputs. 
The second aspect ensures that the code performs well under edge scenarios, which are critical for evaluating the code’s behavior under extreme or unusual conditions. 
These tests are designed to challenge the code with boundary conditions, unexpected inputs, and rare scenarios, to help in identifying potential bugs or weaknesses in the code that might not be evident during basic testing, such as using an empty list or a list with extremely large numbers to test the sorting algorithm.  
Finally, the test designer agent will also generate test cases with large scale values to assess the code’s performance and scalability,, such as testing the sorting algorithm with a list of millions of elements. This involves testing the code under high-load conditions to evaluate whether it maintains its functionality and performance. 
Different from existing methods, 
\xxx generates tests independently without seeing the whole code snippet to keep objectivity and avoid being biased and affected by the incorrect code. The test accuracy and adequacy are compared in Section~\ref{sec:rq3results} and Section~\ref{sec:rq4results}.

\subsection{Test executor agent: code validation and feedback Integration}

Distinct from the programmer agent and test designer agent that are powered by LLMs, the test executor agent in our framework is implemented through a Python script interacting with a local environment and the other two agents (an example of the test executor agent is shown in Appendix Figure 10). As illustrated in~\cref{fig:pipeline}, the test executor agent plays a pivotal role in the final stage of the code generation process. Upon receiving code snippets generated by the programmer agent and test cases generated by the test designer agent, the test executor agent validates these code snippets along with the test cases in a local environment. 
The test executor agent closely monitors the return information from the execution environment (i.e., the terminal). This involves analyzing the output and determining whether the code snippets successfully pass all the test cases. If all test cases are passed, it returns the code to the human developer. Otherwise, if the execution results contain error information~(e.g., syntax errors), the test executor agent will then return the error information to the programmer agent to fix the reported error.

\begin{table*}\scriptsize
    \centering
    \caption{End-to-end results of \xxx and baseline approaches for HumanEval, HumanEval-ET, MBPP, and MBPP-ET datasets. The best approach is highlighted in bold. The baseline results are obtained from its paper report. 
    We use ``-'' to indicate the cases where the results are absent. The percentages in brackets are the improvement rate over the base LLMs (zero-shot prompting). For the last three rows, no baseline optimisation approaches report effectiveness on these LLMs, therefore, we report the results of \xxx only. }
    \begin{tabular}{lrrrrr}
    \toprule
    Models&\textbf{HumanEval}&\textbf{HumanEval-ET}&\textbf{MBPP}&\textbf{MBPP-ET}&\textbf{Mean}\\
    \midrule
    \multicolumn{6}{l}{Zero-Shot LLMs}\\
    \midrule
    AlphaCode~(1.1B)&17.1&-&-&-&17.1  \\
    Incoder~(6.7B)&15.2&11.6&17.6&14.3&14.7 \\
    CodeGeeX~(13B)&18.9&15.2&26.9&20.4&20.4\\
    StarCoder~(15.5B)&34.1&25.6&43.6&33.4&34.2\\
    CodeLlama~(34B)&51.8&-&69.3&-&60.6\\
    Llama3 (8B)&62.2&-&-&-&-\\
    CodeGen-Mono~(16.1B)&32.9&25.0&38.6&31.6&32.0\\
    CodeX~(175B)&47.0&31.7&58.1&38.8&43.9\\
    CodeX~(175B)+CodeT&65.8&51.7&67.7&45.1&57.6\\
    GPT-3.5-turbo&57.3&42.7&52.2&36.8&47.3\\
    PaLM Coder&43.9&36.6&32.3&27.2&35.0\\
    Claude-instant-1&31.1&28.1&26.9&19.9&26.5\\
    GPT-4-turbo&57.9&48.8&63.4&47.5&54.4\\
    GPT-4&67.6&50.6&68.3&52.2&59.7\\
     \midrule
     \multicolumn{6}{l}{LLM-based optimisation methods with GPT-3.5-turbo}\\
     \midrule
Few-Shot& 67.7 (18.2\%)&54.9 (28.6\%)&65.8 (26.1\%)&48.3 (31.2\%)&59.2 (25.2\%)\\
CoT& 44.6 (-22.2\%)&37.2 (-12.9\%)&46.1 (-11.7\%)&34.8 (-5.4\%)&40.7 (-14.0\%)\\
ReAct& 56.9 (-0.7\%)&49.4 (15.7\%)&67.0 (28.4\%)&45.9 (24.7\%)&54.8 (15.9\%)\\
Reflexion& 68.1 (18.8\%)&50.6 (18.5\%)&70.0 (34.1\%)&47.5 (29.1\%)&59.1 (24.9\%)\\
ToT& 54.4 (-5.1\%)&42.7 (0.0\%)&65.8 (26.1\%)&40.8 (10.9\%)&50.9 (7.6\%)\\
RAP& 63.1 (10.1\%)&52.4 (22.7\%)&71.4 (36.8\%)&46.7 (26.9\%)&58.4 (23.5\%)\\
Self-Edit& 62.2 (8.6\%)&54.3 (27.2\%)&56.4 (8.0\%)&45.9 (24.7\%)&54.7 (15.6\%)\\
Self-Planing& 65.2 (13.8\%)&48.8 (14.3\%)&58.6 (12.3\%)&41.5 (12.8\%)&53.5 (13.1\%)\\
Self-debugging& 61.6 (7.5\%)&45.8 (7.3\%)&60.1 (15.1\%)&52.3 (42.1\%)&55.0 (16.3\%)\\
INTERVENOR& 75.6 (31.9\%)&54.8 (28.3\%)&69.8 (33.7\%)&47.1 (28.0\%)&61.8 (30.7\%)\\
CodeCoT& 79.3 (38.4\%)&69.5 (62.8\%)&89.5 (71.5\%)&63.0 (71.2\%)&75.3 (59.2\%)\\

Self-Collaboration& 74.4 (29.8\%)&56.1 (31.4\%)&68.2 (30.7\%)&49.5 (34.5\%)&62.1 (31.3\%)\\
\textbf{\xxx (GPT-3.5-turbo)}&\cellcolor[rgb]{0.9,0.9,0.9}\textbf{ 79.9 (39.4\%)}&\cellcolor[rgb]{0.9,0.9,0.9}\textbf{77.4 (81.3\%)}&\cellcolor[rgb]{0.9,0.9,0.9}\textbf{89.9 (72.2\%)}&\cellcolor[rgb]{0.9,0.9,0.9}\textbf{89.1 (142.1\%)}&\cellcolor[rgb]{0.9,0.9,0.9}\textbf{84.1 (77.8\%)}\\
\midrule
\multicolumn{6}{l}{LLM-based optimisation methods with GPT-4}\\
\midrule
Reflexion& 91.0 (34.6\%)&-&77.1 (12.9\%)&-&84.1 (40.9\%)\\
Self-Debugging& -&-&80.6 (18.0\%)&-&80.6 (35.0\%)\\
Self-Collaboration& 90.2 (33.4\%)&70.7 (39.7\%)&78.9 (15.5\%)&62.1 (19.0\%)&75.5 (26.5\%)\\
ChatDev&84.1 (24.4\%)&-&79.8 (12.9\%)&-&84.1 (40.9\%) \\
AgentVerse& 89.0 (24.4\%)&-&73.5 (7.6\%)&-&81.3 (19.6\%)\\
MetaGPT& 85.9 (27.1\%)&-&87.7 (28.4\%)&-&86.8 (45.4\%)\\
\textbf{\xxx (GPT-4)}&\cellcolor[rgb]{0.9,0.9,0.9} \textbf{96.3 (42.5\%)}&\cellcolor[rgb]{0.9,0.9,0.9}\textbf{86.0 (70.0\%)}&\cellcolor[rgb]{0.9,0.9,0.9}\textbf{91.8 (34.4\%)}&\cellcolor[rgb]{0.9,0.9,0.9}\textbf{91.8 (75.9\%)}&\cellcolor[rgb]{0.9,0.9,0.9}\textbf{91.5 (53.3\%)}\\
\midrule
\multicolumn{6}{l}{LLM-based optimisation methods with other backbone LLMs}\\
\midrule
\textbf{\xxx (PaLM Coder)}&\cellcolor[rgb]{0.9,0.9,0.9} \textbf{64.0 (45.8\%)}&\cellcolor[rgb]{0.9,0.9,0.9}\textbf{55.5 (51.6\%)}&\cellcolor[rgb]{0.9,0.9,0.9}\textbf{75.9 (135.0\%)}&\cellcolor[rgb]{0.9,0.9,0.9}\textbf{75.5 (177.6\%)}&\cellcolor[rgb]{0.9,0.9,0.9}\textbf{67.7 (93.4\%)}\\
\textbf{\xxx (Claude-instant-1)}&\cellcolor[rgb]{0.9,0.9,0.9}\textbf{67.7 (117.7\%)}&\cellcolor[rgb]{0.9,0.9,0.9}\textbf{57.9 (106.0\%)}&\cellcolor[rgb]{0.9,0.9,0.9}\textbf{76.3 (183.6\%)}&\cellcolor[rgb]{0.9,0.9,0.9}\textbf{75.1 (277.4\%)}&\cellcolor[rgb]{0.9,0.9,0.9}\textbf{69.3 (161.5\%)}\\
\textbf{\xxx (GPT-4-turbo)}& \cellcolor[rgb]{0.9,0.9,0.9}\textbf{89.6 (54.7\%)}&\cellcolor[rgb]{0.9,0.9,0.9}\textbf{76.2 (56.1\%)}&\cellcolor[rgb]{0.9,0.9,0.9}\textbf{91.4 (44.2\%)}&\cellcolor[rgb]{0.9,0.9,0.9}\textbf{91.4 (92.4\%)}&\cellcolor[rgb]{0.9,0.9,0.9}\textbf{87.2 (60.3\%)}\\
     \bottomrule
    \end{tabular}
    \vspace{-0.3cm}
    \label{tab:end2end}
\end{table*}

\section{Evaluation}

\subsection{Experiment Setup}
\label{sec:experimentalsetup}
We use pass@1 as the evaluation metric for code correctness, the most widely adopted metric in the literature of automatic code generation~\cite{Austin2021ProgramSW,Chen2021EvaluatingLL,Dong2023CodeScoreEC,Zhang2023SelfEditFC,Dong2023SelfcollaborationCG}.
\paragraph{Datasets.} In this paper, we evaluate \xxx's effectiveness with four widely used code generation datasets, i.e., HumanEval~\cite{chen2021evaluating} and MBPP~\cite{Austin2021ProgramSW}, and their enhanced versions, i.e., HumanEval-ET and MBPP-ET~\cite{Dong2023CodeScoreEC}. HumanEval and HumanEval-ET focus on a range of programming challenges, offering a diverse set of problems to test the model's problem-solving skills and adaptability. On the other hand, MBPP and MBPP-ET provide a comprehensive collection of Python programming problems, designed to evaluate the model's proficiency in Python syntax and its ability to handle a variety of coding scenarios. The enhanced versions, HumanEval-ET and MBPP-ET, include more adequate test cases, making them more challenging and better suited for evaluating advanced models.
We study the effectiveness of \xxx powered by five state-of-the-art LLMs, including GPT-4, GPT-4-turbo, GPT-3.5-turbo, PaLM Coder, and Claude~(Claude-instant-1).

\paragraph{Baselines.} To illustrate the effectiveness of \xxx, we compare \xxx with 12 Large Language Models (LLMs), including open-source and closed-source ones, such as AlphaCode~\cite{Li2022CompetitionlevelCG}, Llama3, CodeLlama~\cite{roziere2023code}, Incoder~\cite{Fried2022InCoderAG}, CodeGeeX~\cite{Zheng2023CodeGeeXAP}, StarCoder~\cite{Li2023StarCoderMT}, CodeGen-Mono~\cite{nijkamp2022codegen}, CodeX~\cite{Brown2020LanguageMA}, GPT-3.5-turbo, and GPT4~\cite{OpenAI2023GPT4TR}. These models vary in architecture, training methodologies, and application scopes. Additionally, we compare \xxx with 16 state-of-the-art (SOTA) code generation methods that are based on LLMs but with various optimisation strategies, including Few-shot learning, Chain-of-Thought~\cite{Wei2022ChainOT}, ReAct~\cite{Yao2022ReActSR}, Reflexion~\cite{Shinn2023ReflexionLA}, ToT~\cite{Yao2023TreeOT}, RAP~\cite{Hao2023ReasoningWL}, Self-Edit~\cite{Zhang2023SelfEditFC}, Self-Planing~\cite{Jiang2023SelfplanningCG}, Self-Debugging~\cite{Chen2023TeachingLL}, Self-Collaboration~\cite{Dong2023SelfcollaborationCG}, SCOT~\cite{Li2023StructuredCP},
CodeCoT~\cite{Huang2023CodeCoTAB}, 
and INTERVENOR~\cite{Wang2023INTERVENORPT}.
These methods have been shown to significantly enhance the performance of LLMs in complex problem-solving scenarios.

\subsection{RQ1: How does \xxx perform?}

As shown in \cref{tab:end2end}, we can observe that \xxx outpeforms all the base LLM models and all the baseline optimisation approaches in all the datasets. 
Specifically, if we focus on the improvement that \xxx achieves over the base LLMs, take GPT-3.5-turbo as an example, 
GPT-3.5-turbo obtains 57.3\% pass@1 in the HumanEval dataset, while \xxx obtains 79.9\%.
For GPT-4, the mean pass@1 of \xxx is 91.5\% across all the datasets, \textbf{32.7\% improvement} over the baseline zero-shot GPT-4 model.
For PaLM Coder, Claude-instant-1, and GPT-4-turbo, the mean improvement of \xxx over the base models are \textbf{32.7\%, 42.8\%, 32.8\%}, respectively.

\xxx also demonstrates superiority over all optimisation baselines.
For example, for MBPP-ET with GPT-3.5-turbo,
\xxx obtains 89.1\% pass@1, while CodeCoT, the state-of-the-art approach, achieves only 63.0\%. 
On average, the pass@1 of \xxx is 84.1\%, \textbf{8.8\% more than the state-of-the-art approach CodeCoT}.
{\color{black}One reason for \xxx's superiority over CodeCoT is that CodeCoT generates tests and code at the same time with only one agent,}
while \xxx has the test designer agent which generates more powerful test cases. 
RQ4 and RQ5 introduce more analysis on their comparison in terms of the effectiveness of test cases. 

The HumanEval-ET and MBPP-ET datasets contain more comprehensive tests and are more challenging for code generation approaches to get high pass@1.
We can observe that the base LLMs and the baseline optimisation approaches perform significantly worse on these two enhanced versions.
However, \xxx's performance on these enhanced datasets is comparative to the original datasets, which is another superiority of \xxx, largely because the test designer agent generates rigorous tests to ensure that the generated code is indeed reliable.

\begin{table}
    \centering
    \caption{Contribution of different agents in \xxx.}
    \resizebox{0.8\linewidth}{!}{
    \begin{tabular}{lrrrr}
    \toprule
    Agents&HumanEval&HumanEval-ET&MBPP&MBPP-ET\\
    \midrule
    programmer agent only&61.0&52.4&47.9&35.0\\
programmer + test designer& 64.0 (11.7\%)&54.3 (27.2\%)&62.3 (19.3\%)&45.9 (24.7\%)\\
programmer + test executor& 64.6 (12.7\%)&55.5 (30.0\%)&69.3 (32.8\%)&51.4 (39.7\%)\\
    \xxx&\textbf{79.9 (39.4\%)}&\textbf{77.4 (81.3\%)}&\textbf{89.9 (72.2\%)}&\textbf{89.1 (142.1\%)}\\
         \bottomrule
    \end{tabular}}
    \label{tab:component}
\end{table}

\subsection{RQ2: How do different agents contribute to the effectiveness of \xxx?}\label{sec:eval:end2end}
As illustrated in~\cref{fig:pipeline}, \xxx contains three agents, i.e., the programmer agent, the test designer agent, and the test executor agent, where the programmer agent focuses on generating code snippets based on the code generation requirements and feedback from other agents. The test designer agent focuses on generating test cases, which are used to evaluate the correctness of the code snippets produced by the programmer agent. The test executor agent interacts with the other two agents to collect the code snippets and test cases and executes them in a local environment to prepare feedback. This research question investigates how each agent contributes to \xxx's effectiveness with four agent combination scenarios, i.e., the programmer agent itself, the programmer + test designer agent, where we feed the function and test cases into the programmer agent and require it to analyze whether it needs to refine the code to pass all test cases, and the programmer + test executor agent, where we directly run the generated code with the tests provided in the prompt \footnote{The code generation prompts in HumanEval and MBPP contain a few test cases.
}(we provide the programmer + test designer/executor agent prompts in Appendix Figure 11 and 12).

The evaluation results are shown in~\cref{tab:component}. We can observe that first, with the assistant of the test designer and the test executor agent, the pass@1 increases compared with the result of only the programmer agent. For example, with both the programmer and the test designer agent, the pass@1 increases from 61.0\% to 64.0\%. However, without the test executor agent, the programmer agent is not able to get reliable feedback from dynamic test case execution. Therefore, the performance is significantly below \xxx. 
For the programer + test executor agent, it obtains 64.6\% and 69.3\% pass@1 in HumanEval and MBPP, which is also higher than the programmer agent itself which obtains 61.0\% and 47.9\%. This is because test executor agent detects some bugs in the code with the test cases provided by the prompt.
However, the number of test cases is very limited, with only two to three tests in HumanEval and MBPP.
The effectiveness of these tests are far below from the tests generated by the test designer agent.
Therefore, without the test designer agent, the performance is also significantly below \xxx.

\subsection{RQ3: How do code refinement iterations affect \xxx's effectiveness?}
\label{sec:rq3results}
\xxx refines code snippets based on the feedback provided by the test executor agent. In this experiment, we evaluate how the number of refinement iterations affect \xxx's effectiveness. Specifically, we analyze \xxx's effectiveness with its result for each refinement iteration. Table~\ref{tab:steps} shows the results, we can observe that the pass@1 increase with more iterations. In particular, when we increase the number of iterations from 1 to 5, the pass@1 of HumanEval and HumanEval-ET increases from 74.4\% to 79.9\% and 73.2\% to 77.4\%. We can also observe these behaviors for the MBPP and MBPP-ET datasets, where the pass@1 increases from 84.1\% to 89.9\% and 80.3\% to 89.1\%. 

\begin{table}
    \centering
    \caption{Pass@1 of \xxx with different number of iterations on GPT-3.5-turbo.}
    \resizebox{0.7\linewidth}{!}{
    \begin{tabular}{c|cccc}
    \toprule
    Iterations&HumanEval&HumanEval-ET&MBPP&MBPP-ET\\
    \midrule
1& 74.4 (29.8\%)&73.2 (71.4\%)&84.1 (61.1\%)&80.3 (118.2\%)\\
2& 75.6 (31.9\%)&73.2 (71.4\%)&86.4 (65.5\%)&85.6 (132.6\%)\\
3& 76.2 (33.0\%)&75.0 (75.6\%)&87.9 (68.4\%)&87.6 (138.0\%)\\
4& 78.7 (37.3\%)&76.8 (79.9\%)&88.7 (69.9\%)&88.7 (141.0\%)\\
5& 79.9 (39.4\%)&77.4 (81.3\%)&89.9 (72.2\%)&89.1 (142.1\%)\\
         \bottomrule
    \end{tabular}
    }
    \label{tab:steps}
\end{table}

\subsection{RQ4: How accurate are the tests generated by the test designer agent?}
\label{sec:rq4results}

\begin{wraptable}{r}{0.4\textwidth}\scriptsize
    \centering
    \vspace{-0.5cm}
    \caption{Accuracy of the test cases.}
    \begin{tabular}{lrr}
    \toprule
    Models&HumanEval&MBPP\\
    \midrule
        GPT-3.5-turbo&47.0&57.2\\
        CodeCoT&67.1&79.0\\
         \xxx (GPT-3.5-turbo)&87.8&89.9\\
         \hline
         MetaGPT (GPT-4)&79.3&84.4\\
         \xxx (GPT-4) &89.6&91.4\\
         
         \bottomrule
    \end{tabular}
    \label{tab:testcase}
\end{wraptable}
The test designer agent focuses on generating test cases to analyze whether the code has bugs and plays a crucial role in \xxx. However, once the test cases are incorrect (e.g., with incorrect test oracles), the feedback the test cases provide will be problematic, misleading the programmer agent and decreasing \xxx's overall effectiveness. 
Therefore, this research question investigates how reliable the test designer agent is in generating accurate tests to aid the programmer agent. 
We evaluate the accuracy of the test cases under the datasets' \textbf{canonical solution}\footnote{Each coding task in the datasets has a canonical solution, which is the ground truth for code generation.} on GPT-3.5-turbo and GPT-4. The tests that pass the canonical solution are correct. 
To demonstrate the effectiveness of the test designer agent in \xxx, we compare the accuracy of the tests generated by \xxx, the GPT-3.5-turbo, CodeCoT, and MetaGPT, where the tests are generated at the same time with the code in a non-independent way.

The evaluation results are shown in~\cref{tab:testcase}.
First, we observe that the accuracy of the tests cases produced by the test designer agent in \xxx is 87.8\% and 89.9\%, respectively in HumanEval and MBPP datasets for GPT-3.5-turbo backbone, while GPT-3.5-turbo obtains only 47.0\% and 57.2\%. In addition, we observe that the test designer agent in \xxx (GPT-4) is also more accurate than MetaGPT (GPT-4) in test generation. For example, on HumanEval, the accuracy is 89.6\% v.s. 79.3\% for \xxx and MetaGPT. The superiority of \xxx demonstrates the effectiveness of the prompt engineering strategies we designed for the test designer agent. 

\subsection{RQ5: How adequate are \xxx's test cases in code coverage?}

\begin{wraptable}{r}{0.4\textwidth}\scriptsize
    \centering
    \vspace{-0.65cm}
    \caption{Line coverage of the tests.}
    \begin{tabular}{lrr}
    \toprule
    Models&HumanEval&MBPP\\
    \midrule
    GPT-3.5-turbo&70.2&61.3\\
    CodeCoT&77.2&82.9\\
 \xxx (GPT-3.5-turbo)&87.5&89.5\\
 \hline
 MetaGPT (GPT-4) &81.7&80.5\\
 \xxx (GPT-4)&91.7&92.3\\
         \bottomrule
    \end{tabular}
    \label{tab:coverage}
\end{wraptable}
This research question explores the adequacy of the test cases generated by the test designer agent in code coverage. Specifically, we evaluate how many lines of code in the canonical solution are covered by the test cases generated by the original GPT-3.5-turbo, CodeCoT, MetaGPT, and \xxx. The evaluation results were illustrated in~\cref{tab:coverage}, where we can observe that the tests generated by \xxx have the highest code coverage. For example, \xxx (GPT-3.5-turbo) obtains 87.5\% and 89.5\% code coverage compared with CodeCoT (GPT-3.5-turbo), which only obtains 77.2\% and 82.9\%, on the two datasets when we calculate the code line coverage with the all tests generated by each strategy. Besides, \xxx (GPT-4) also obtains 91.7\% code line coverage in the HumanEval dataset, while MetaGPT (GPT-4) only obtains 81.7\% code line coverage.

\subsection{RQ6: Should programmer and test designer be separated to different agents?}

\begin{wraptable}{r}{0.4\textwidth}\scriptsize
    \centering
    \caption{Accuracy of the tests generated by single- and multi-agents.}
    \begin{tabular}{c|cc}
    \toprule
    Models&HumanEval&MBPP\\
    \midrule
         Single Agent&61.0&51.8\\
         Multiple Agents&87.8&89.9 \\
         \bottomrule
    \end{tabular}
    \label{tab:single_agent_tests}
\end{wraptable}

\xxx requires separate agents for generating code and tests (i.e., the programmer and test designer agent). An alternative way is to let a single agent first generate code and then generate tests, within the same conversation.
This research question investigates 
whether requiring one agent to finish two tasks, i.e., code generation and test case generation, is as effective as using separate agents. 

\begin{table}
    \centering
    \caption{Pass@1 for a single agent and multiple agents.}
    \resizebox{0.7\columnwidth}{!}{
    \begin{tabular}{c|cccc}
    \toprule
         Models&HumanEval&HumanEval-ET&MBPP&MBPP-ET  \\
         \midrule
         Single Agent&71.3&61.6&79.4&59.1 \\
         Multiple Agents&\textbf{79.9}&\textbf{77.4}&\textbf{89.9}&\textbf{89.1}\\
         \bottomrule
    \end{tabular}}
    \label{tab:single_agent_code}
    \vspace{-0.3cm}
\end{table}

The evaluation results are shown in~\cref{tab:single_agent_code}, \cref{tab:single_agent_tests}, and \cref{tab:rq6_coverage}. We can observe that the pass@1 of using a single agent to generate both code and tests is lower than assigning the two tasks to different agents. For example, the pass@1 of the single agent has only  71.3\% and 79.4\% pass@1 for HumanEval and MBPP, while the multi-agent setup~(\xxx) obtains 79.9\% and 89.9\% for HumanEval and MBPP. We also observe that the test case accuracy for the single agent is also lower than the multi-agent setting~(\xxx). Specifically, the single agent only obtains 61.0\% and 51.8\% in HumanEval and MBPP datasets, while the multi-agent setup~(\xxx) obtains 87.8\% and 89.9\% in HumanEval and MBPP. 
Finally, as shown in \cref{tab:rq6_coverage}, we can also observe that the tests' coverage results of the single agent are also lower than in the multi-agent setup. For example, the single agent only obtains 72.5\% and 75.9\% code line coverage while multiple agents obtain 87.5\% and 89.5\% code line coverage.

\begin{wraptable}{r}{0.4\textwidth}\scriptsize
    \centering
    \caption{Code line coverage of tests generated by single agent and multi-agent setup. }
    \begin{tabular}{lrr}
    \toprule
    Models&HumanEval&MBPP\\
    \midrule
        Single Agent&72.5&75.9\\
         Multiple Agents&87.5&89.5\\
         \bottomrule
    \end{tabular}
    \label{tab:rq6_coverage}
\end{wraptable}

There are two possible reasons for the superiority of the multi-agent setup. First, letting a single agent do both code generation and test case design may distract the agent's focus; second, the tests designed by the same agent that generates the code can be biased by the code and lose objectivity, for example, if the generated code ignores the handling of edge cases, the generated tests can be affected by flaws in the code.
These results demonstrate the necessity of using multiple agents to collaborate in code generation, with different agents taking different roles.
{\color{black}Such benefit of multi-agent collaborations with LLMs has also been illustrated in other multi-agent systems~\cite{Chen2023AgentVerseFM,Zhang2023ProAgentBP}.}

\section{Conclusion}

In this paper, we have proposed \xxx, which contains multiple agents to improve the code generation effectiveness of code generation models with effective and accurate automated test generation. \xxx contains three agents, i.e., the programmer, the test designer, and the test executor agent. 
Throughout our evaluations, \xxx demonstrated state-of-the-art performance, outperforming existing LLMs and prompt engineering methods in a variety of coding scenarios. For example, GPT-4 achieves a pass@1 rate of 96.3\% on the HumanEval dataset and 91.8\% on the MBPP dataset, with a token overhead of 56.9K and 66.3K, respectively. In contrast, the current state-of-the-art models achieve a pass@1 rate of 90.2\% and 78.9\% on these datasets, with significantly higher token overheads of 138.2K and 206.5K, respectively.
The limitations and broader impact are discussed in the Appendix. 

\bibliographystyle{named}
\bibliography{ijcai24}

\newpage
\clearpage
\appendix
\appendix
\section{Appendix}\label{sec:appendix}

\subsection{Response Setup}
To ensure that the output of each agent follows our requirements for the execution of the test executor agent, we will require each agent's output follow the architecture of \texttt{```py[Code]'''} and \texttt{```py[TestCases]'''}, where the [Code] and [TestCases] will be in the \texttt{```py'''}. With this format, the test executor agent can directly obtain [Code] and [TestCases] by removing the other sentences before and after these code blocks, ensuring an accurate and focused analysis.

\subsection{Overhead}
In \cref{tab:end2end}, we only discuss the pass@1 of AgentCoder and simultaneous multi-agent works. In this section, we further discuss the overhead of \xxx and these baselines. The evaluation results are demonstrated in \cref{tab:comparison}, where we can observe that AgentCoder requires lower tokens and execution time compared with baseline multi-agent frameworks.

\begin{table}[h!]
    \centering
    \caption{pass@1 of AgentCoder and baselines in GPT4. We utilize the tiktoken package to calculate agent response token usage. Tokens and Overhead are calculated for the HumanEval / MBPP.}
    \resizebox{\columnwidth}{!}{
    \begin{tabular}{lrrrrr}
    \toprule
        Model &HEval&MBPP&Tokens&Overhead&Status\\
        \midrule
        SelfCollaboration&90.2&78.9&74.3k / 89.2K&249.2 / 395.6&Not Yet \\
        AgentVerse&89.0&73.5&149.2K / 193.6K&1573.2 / 1875.5&ICLR (16-01-2024)\\
         ChatDev&84.1&79.8&183.7K / 259.3K&1925.7 / 2493.4&Not Yet\\
         MetaGPT&85.9& 87.7&138.2K / 206.5K&1248.5 / 1583.6&ICLR (16-01-2024)\\
         \textbf{AgentCoder}&96.3&91.8&56.9K / 66.3K&228.7 / 365.9&{------}\\
         \bottomrule
    \end{tabular}}
    \label{tab:comparison}
\end{table}

\subsection{Case Illustration for CodeCoT and \xxx}
\begin{figure*}
    \centering
    \includegraphics[width=0.8\textwidth]{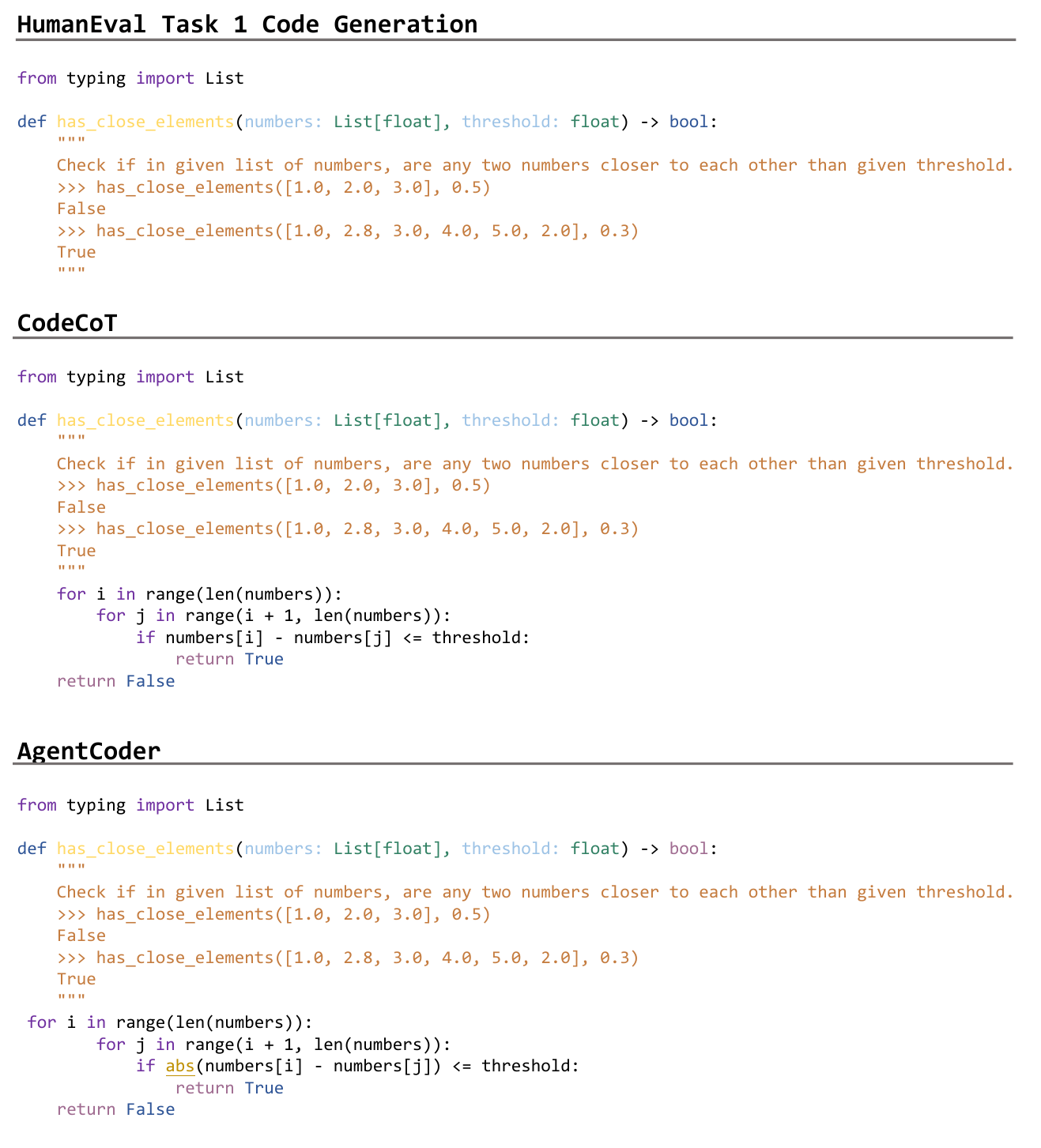}
    \caption{A case illustration of CodeCoT and \xxx generated code for HumanEval task. CodeCoT ignores to use of \textit{abs()} function to check further the absolute values are lower than the threshold, while \xxx employs it to handle the negative values.}
    \label{fig:human_code_example}
\end{figure*}

\begin{figure*}
    \centering
    \includegraphics[width=0.8\textwidth]{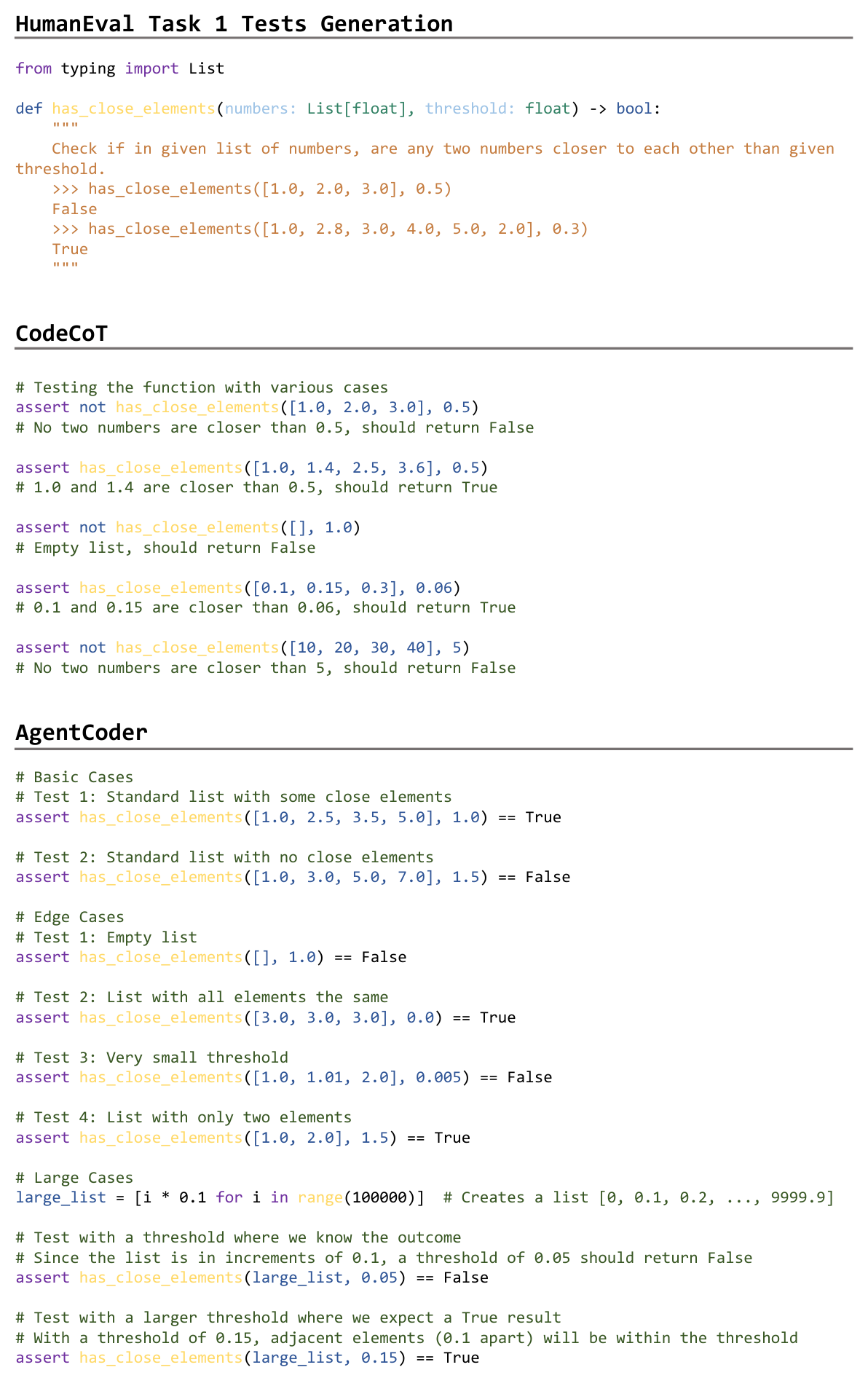}
    \caption{A case illustration of CodeCoT and \xxx generated tests for HumanEval task. CodeCoT only considers the left values to be lower than the right values, which is due to the tests generated with its code where it also ignores the use of the \textit{abs()} function, while \xxx considers two scenarios~(i.e., left value lower/larger than the right values).}
    \label{fig:human_test_example}
\end{figure*}

\begin{figure*}
    \centering
    \includegraphics[width=0.8\textwidth]{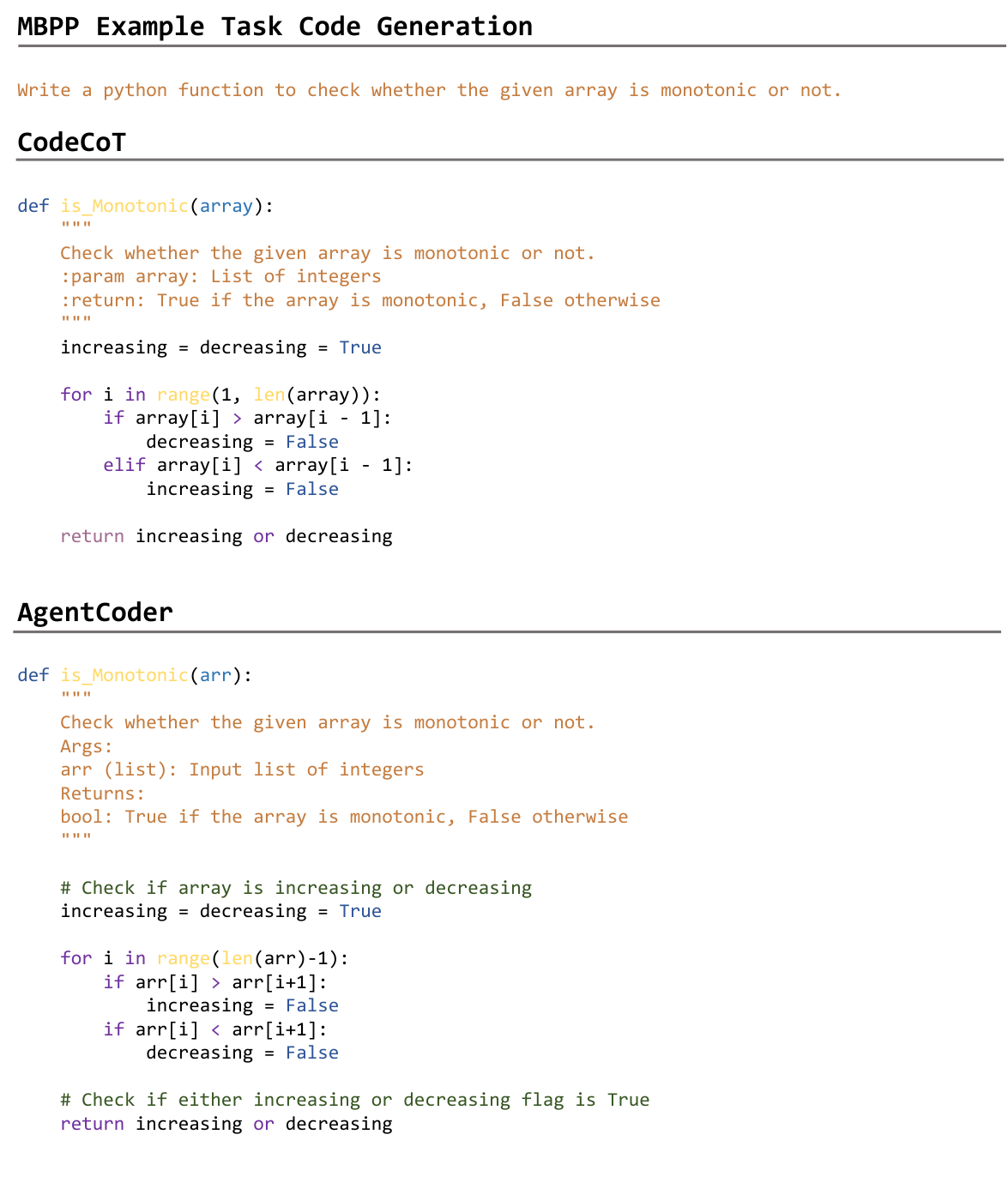}
    \caption{A case illustration of CodeCoT and \xxx generated code for MBPP task. Both CodeCoT and \xxx's code are correct. However, CodeCoT ignores the edge cases~(e.g., the list does not contain values).}
    \label{fig:mbpp_code_example}
\end{figure*}

\begin{figure*}
    \centering
    \includegraphics[width=0.8\textwidth]{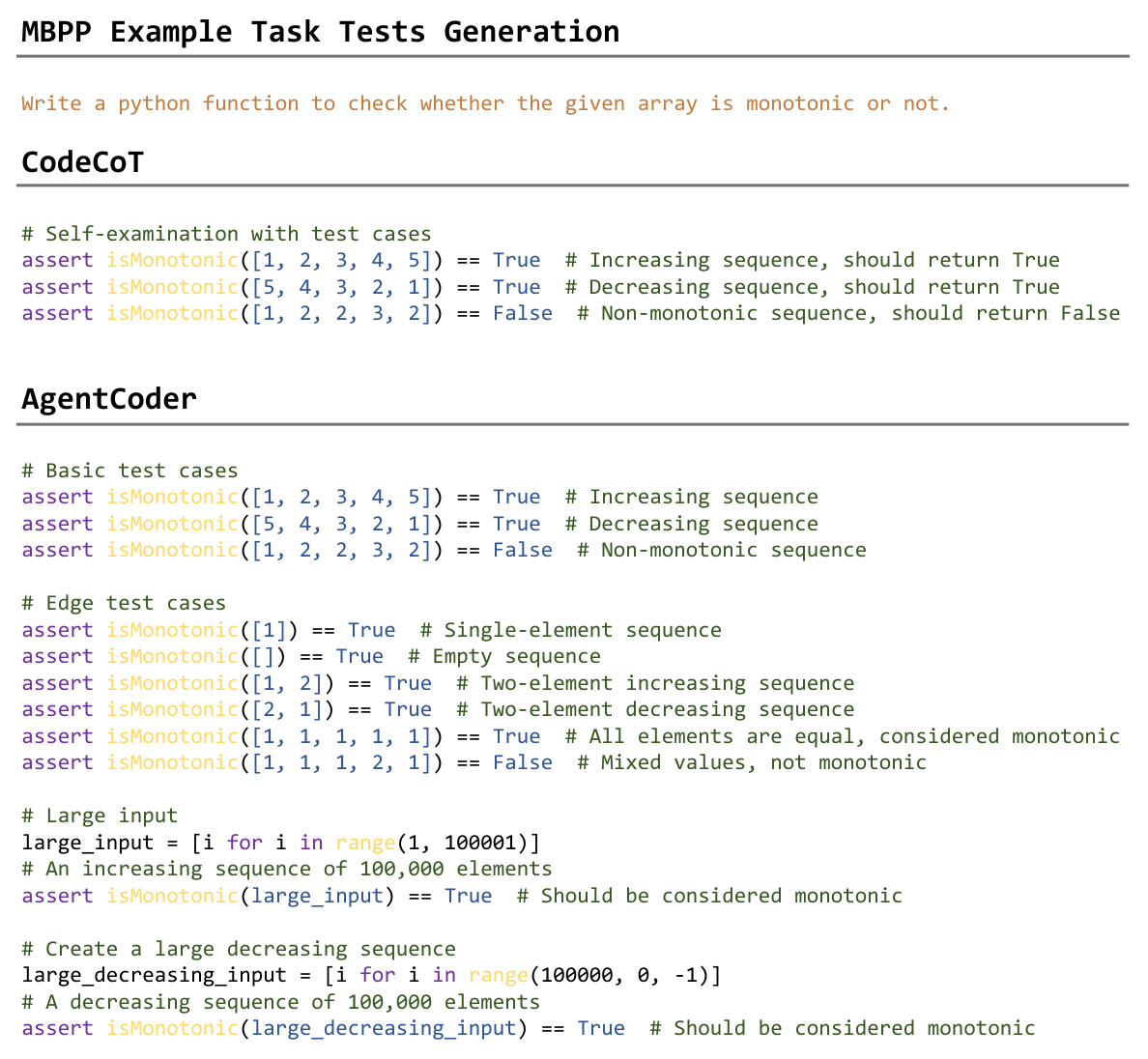}
    \caption{A case illustration of CodeCoT and \xxx generated tests for MBPP task. CodeCoT ignores to consider the list does not contain values and in its generated code this scenario is also ignored. However, \xxx's edge cases will cover these edge scenarios.}
    \label{fig:mbpp_test_example}
\end{figure*}

To provide a comprehensive illustration for CodeCoT and \xxx, we provide two code and tests generation examples 
for HumanEval and MBPP datasets from Fig.~\ref{fig:human_code_example} to Fig.~\ref{fig:mbpp_test_example}. We can observe that \xxx can generate more fine-grained tests for the generated code. For example, \xxx will consider the code execution results when the input list does not contain element~(\cref{fig:human_test_example} and \cref{fig:mbpp_test_example}), which can improve code snippet reliability for edge behaviors.

\subsection{Case Illustration on HumanEval dataset using \xxx}
\begin{figure*}[h]
    \centering
    \includegraphics{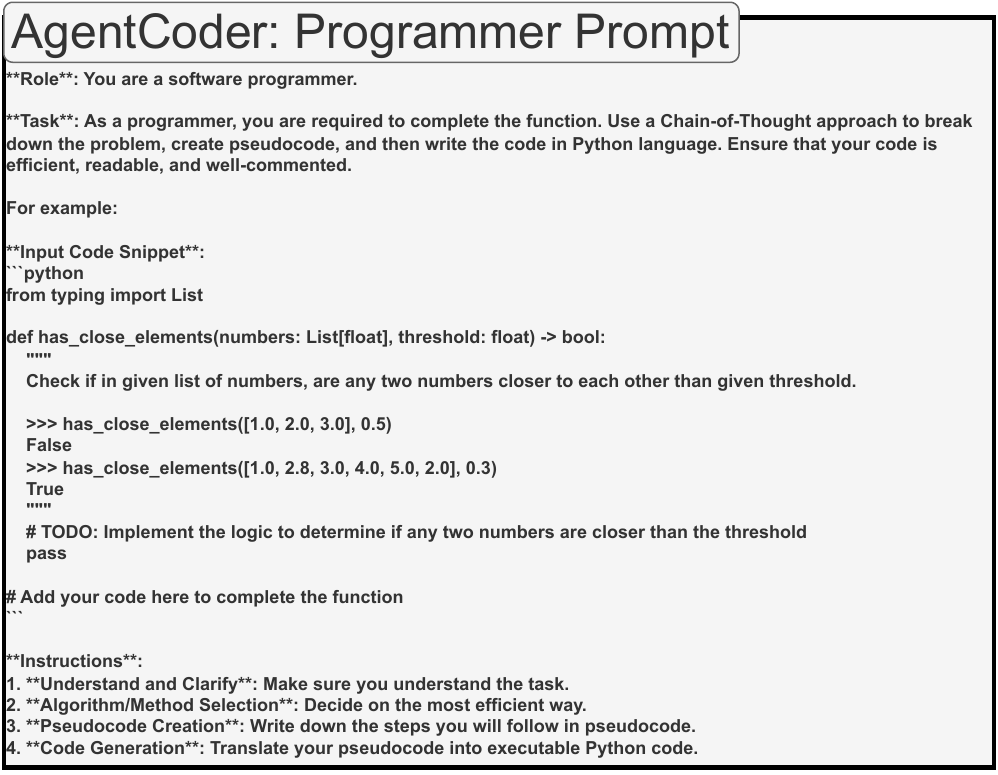}
    \caption{\xxx programmer prompt example.}
    \label{fig:programmer_prompt}
\end{figure*}

\begin{figure*}[h]
    \centering
    \includegraphics[width=0.8\textwidth]{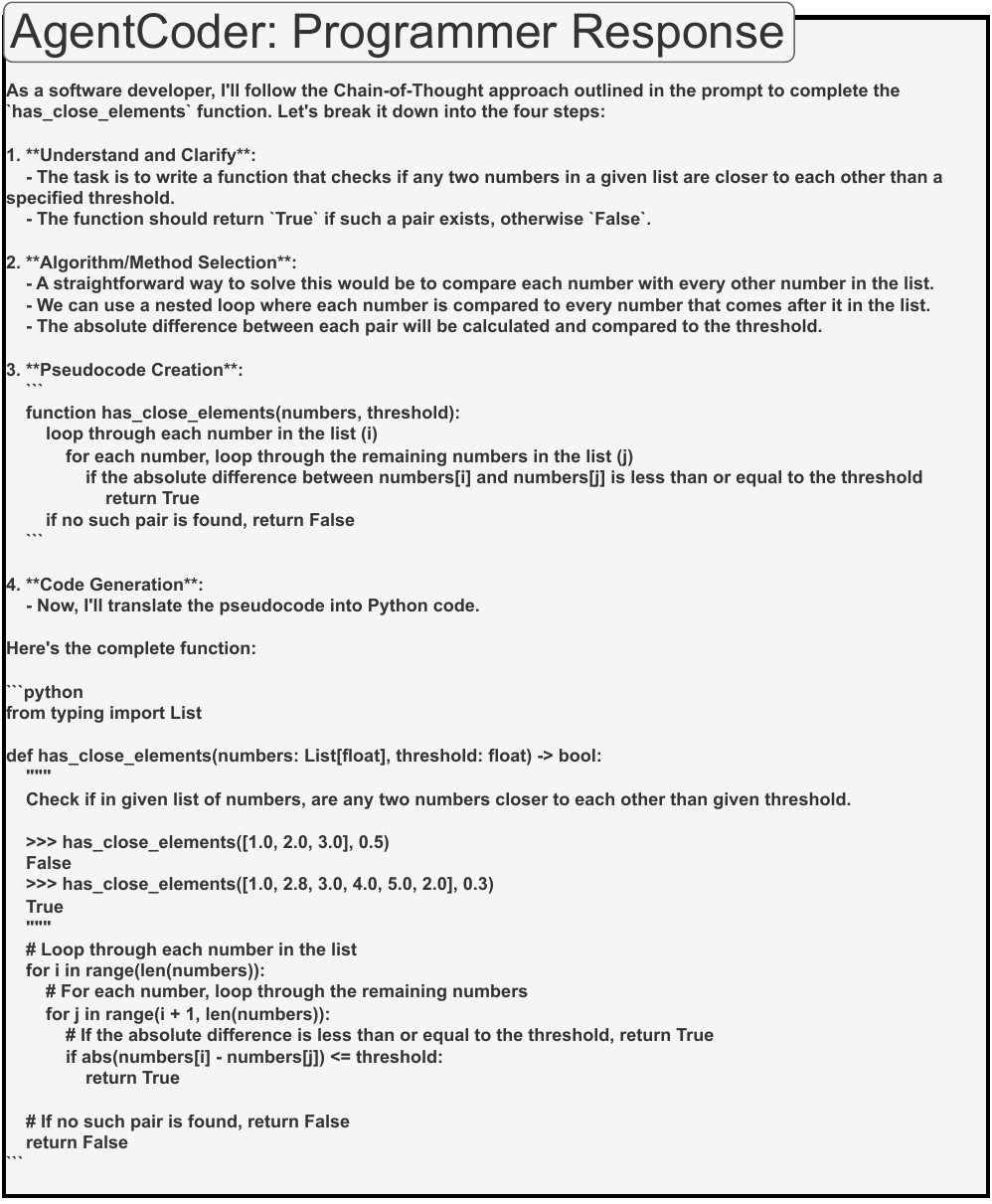}
    \caption{\xxx programmer response example.}
    \label{fig:programmer_response}
\end{figure*}

\begin{figure*}[h]
    \centering
    \includegraphics[width=0.8\textwidth]{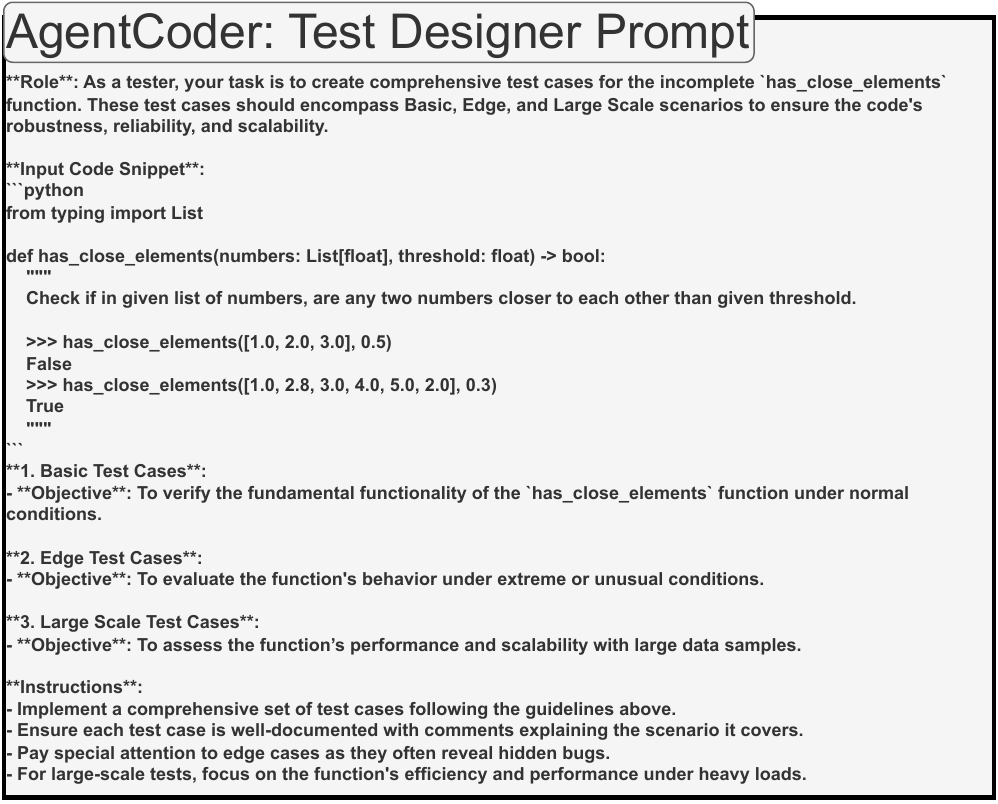}
    \caption{\xxx tester prompt example.}
    \label{fig:test_designer_prompt}
\end{figure*}

\begin{figure*}[h]
    \centering
    \includegraphics[width=0.8\textwidth]{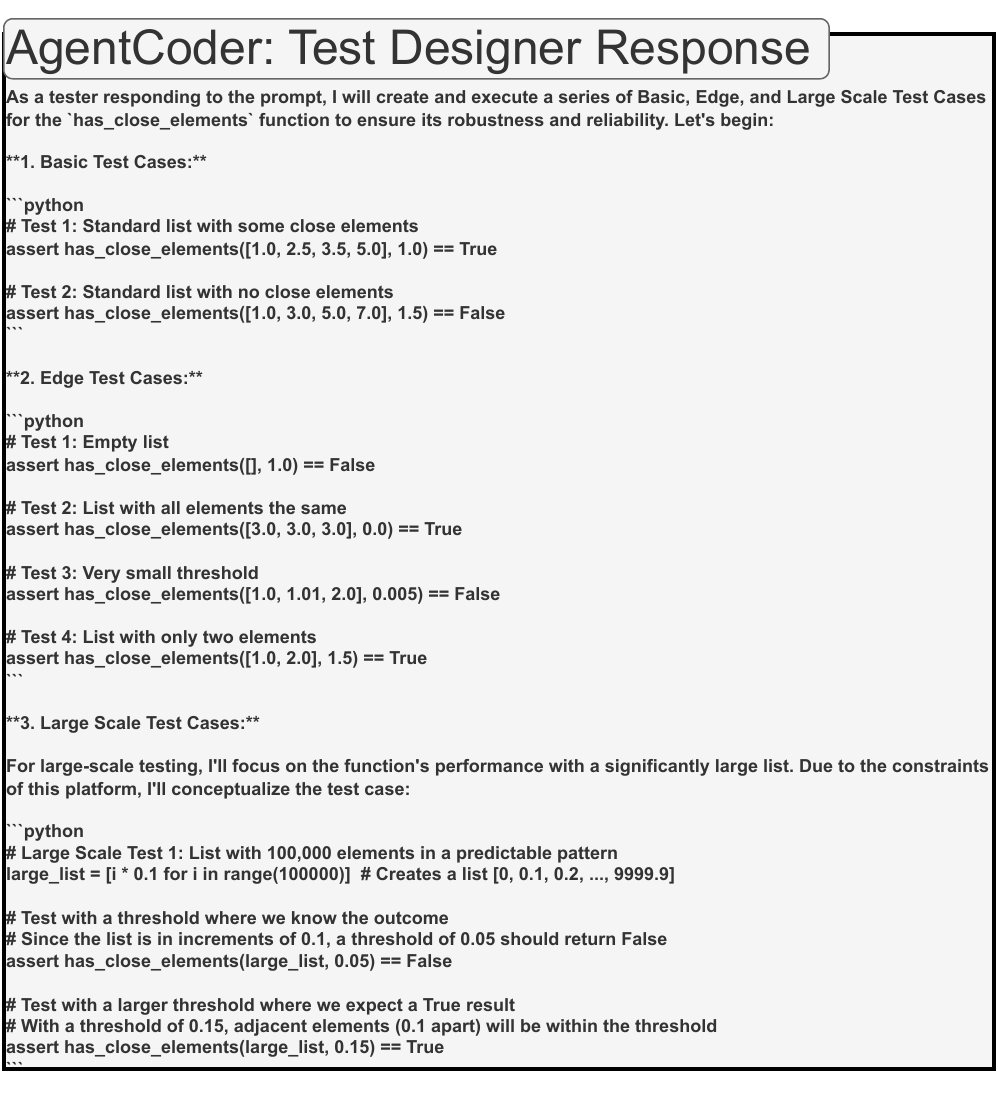}
    \caption{\xxx test designer response example.}
    \label{fig:test_designer_response}
\end{figure*}

\begin{figure*}[h]
    \centering
    \includegraphics[width=0.8\textwidth]{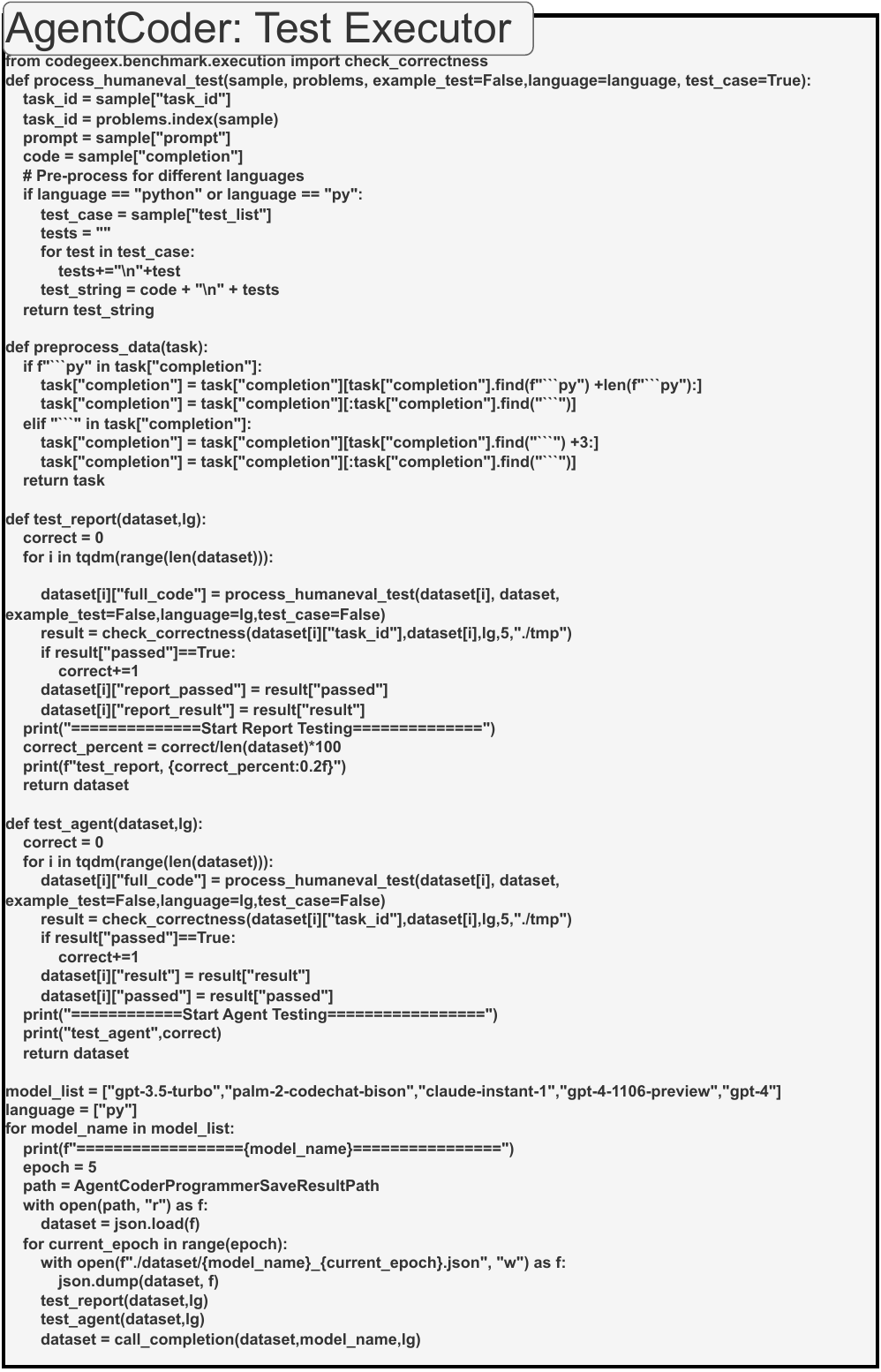}
    \caption{\xxx test executor script.}
    \label{fig:executor}
\end{figure*}

We also provide each agent's prompt and response example~(\cref{fig:programmer_prompt} to \cref{fig:executor}) to illustrate \xxx's workflow. \cref{fig:programmer_prompt} and \cref{fig:programmer_response} illustrate \xxx's programmer prompt and response example. \cref{fig:test_designer_prompt} and \cref{fig:test_designer_response} provide \xxx's test designer prompt and response example. \cref{fig:executor} illustrates \xxx's test executor source code.

\subsection{Case Illustration of the programmer + test executor agent}
\begin{figure*}[h]
    \centering
    \includegraphics[width=0.8\textwidth]{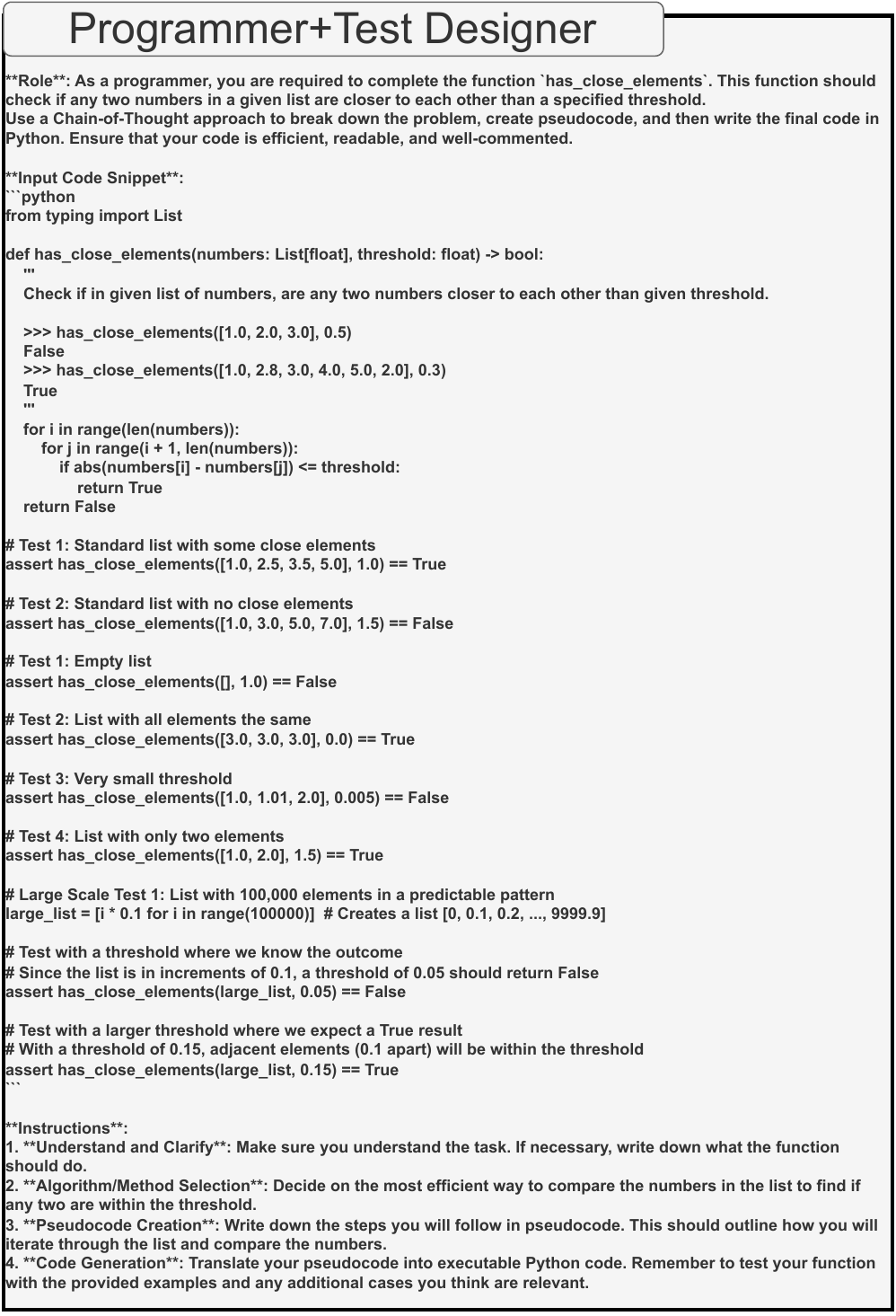}
    \caption{Programmer + test designer example.}
    \label{fig:programmer_executor}
\end{figure*}

We illustrate the pipeline of the programmer + the test executor agent in~\cref{fig:programmer_executor}.

\subsection{Case Illustration of the programmer + test designer}
\begin{figure*}[h]
    \centering
    \includegraphics[width=0.8\textwidth]{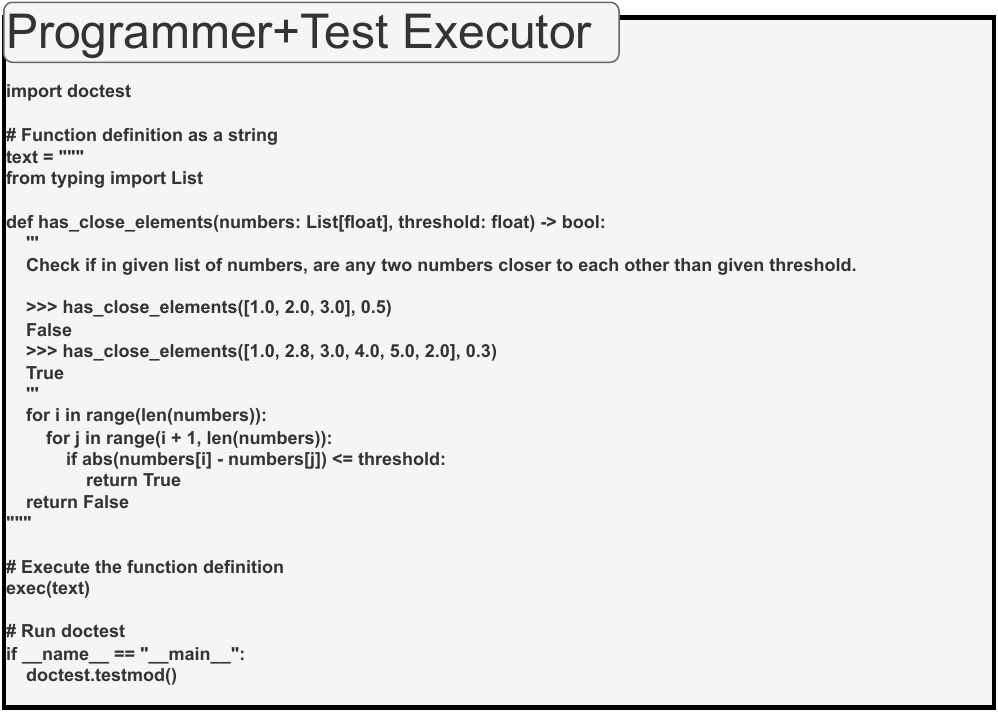}
    \caption{Programmer + test executor example.}
    \label{fig:programmer_designer}
\end{figure*}

We illustrate the pipeline of the programmer + the test designer agent in~\cref{fig:programmer_designer}.

\end{document}